\documentclass[11pt]{article}

\usepackage{acl}
\usepackage{times}
\usepackage{latexsym}
\usepackage[T1]{fontenc}
\usepackage[utf8]{inputenc}
\usepackage{microtype}
\usepackage{inconsolata}
\usepackage{graphicx}
\usepackage{booktabs}
\usepackage{multirow}
\usepackage{url}

\title{LegalCiteTrust: Benchmarking Citation Trustworthiness in Chinese Long-Form Legal Research Reports}

\author{
Yunhan Li\textsuperscript{1,2} \quad
Mingjie Xie\textsuperscript{3} \quad
Zeyang Shi\textsuperscript{1} \quad
Gengshen Wu\textsuperscript{1,*} \quad
Min Yang\textsuperscript{2,*}\\
\textsuperscript{1}Faculty of Data Science, City University of Macau, Macau\\
\textsuperscript{2}Shenzhen Institutes of Advanced Technology, Chinese Academy of Sciences, China\\
\textsuperscript{3}Southern University of Science and Technology, China\\
\texttt{D24092110205@cityu.edu.mo},
\texttt{gswu@cityu.edu.mo},
\texttt{min.yang@siat.ac.cn}\\
\textsuperscript{*}Corresponding author.
}

\begin{document}
\maketitle
\thispagestyle{plain}
\pagestyle{plain}
\begin{abstract}
Long-form legal research reports increasingly rely on LLMs and agentic research systems, but their reliability depends not only on answering the task, but also on whether cited legal authorities are trustworthy. A citation can be risky even when it points to a real source: the report may omit limiting conditions, misdescribe the authority, or use it to support a stronger claim than the source allows. We introduce LegalCiteTrust, a benchmark for evaluating citation trustworthiness in Chinese long-form legal research reports. It contains 72 densely annotated report-level tasks and evaluates reports along three dimensions: Coverage, Support, and Citation Trustworthiness. Citation Trustworthiness is operationalized through citation-level Existence, Fidelity, and Applicability (E/F/A). Experiments on general-purpose LLMs, deep-research systems, and legal-specific systems show that task completion, evidence richness, citation density, and citation reliability expose different system behaviors. Retrieval tools can improve evidence support without reliably improving the Trust score, while E/F/A-based revision improves Trust and Final score more clearly than existence-only filtering. These results suggest that trustworthy legal research generation requires citation-aware evidence governance after retrieval: systems must not only retrieve legal authorities, but also select, describe, and apply them reliably.
\end{abstract}

\section{Introduction}
Long-form legal research reports differ from short legal answers because they must organize issues across subtopics and ground the analysis in statutes, judicial interpretations, regulations, and cases. As LLMs and agentic research systems begin to automate this form of report writing~\cite{avraham2026dream}, citations become the legal authorities on which the report appears to rest. The central risk is therefore not only whether a system reaches a plausible conclusion, but whether the citations supporting that conclusion are trustworthy. In this paper, citation trustworthiness means that a cited legal authority exists, is faithfully described, and supports the local claim for which it is used. A nonexistent, distorted, or misapplied citation can make a report appear legally grounded while resting on unreliable authority~\cite{legg2025promise}.

Existing evaluations cover important parts of this problem, but not this relation in full. Legal benchmarks and practice-oriented evaluations measure legal knowledge, reasoning, retrieval, or task completion~\cite{fei2024lawbench,shi2026plawbench}; citation-oriented work more directly evaluates whether citations can be generated, matched, or verified~\cite{chen2026legalcitebench}. What remains under-evaluated is not simply whether a report is useful, complete, or citation-rich, but whether each cited legal authority is used as a trustworthy ground for the local claim. This matters because citation failures are contextual: a model may cite a real source while omitting a limiting condition, or accurately describe a rule while using it to support a stronger claim than the source allows.

\begin{figure*}[t]
\centering
\includegraphics[width=0.85\textwidth]{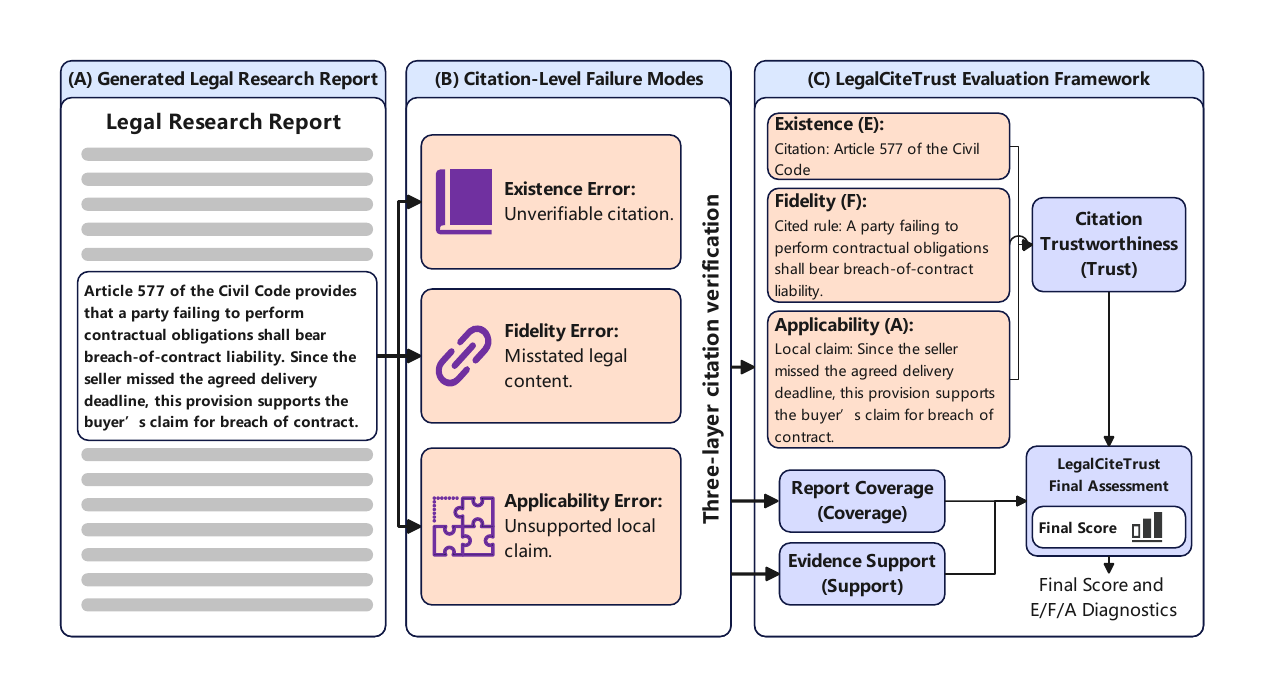}
\caption{Overview of LegalCiteTrust. The benchmark connects report-level Coverage and Support with citation-level Existence, Fidelity, and Applicability judgments for evaluating citation trustworthiness.}
\label{fig:overview}
\end{figure*}

We introduce LegalCiteTrust, a benchmark for Chinese long-form legal research reports. It separates three aspects often conflated in evaluation: whether a report follows the requested research structure, whether it provides legal evidence, and whether its citations are trustworthy. Citation trustworthiness is evaluated through Existence, Fidelity, and Applicability (E/F/A). Figure~\ref{fig:overview} summarizes the task setting and evaluation framework.

Using LegalCiteTrust, we evaluate general-purpose LLMs, deep-research systems, and legal-specific systems, with tool and verification ablations. The results show that task completion, evidence richness, citation density, and citation reliability do not collapse into one quality axis: retrieval can increase evidence support without reliably improving citation trustworthiness, while E/F/A feedback improves revision more clearly than existence-only filtering. These findings suggest that trustworthy legal research generation requires citation-aware evidence governance after retrieval, not merely more retrieval.

\section{Related Work}
LegalCiteTrust is positioned at the intersection of legal task evaluation, agentic or deep-research report evaluation, and hallucination or citation-faithfulness evaluation. Table~\ref{tab:related-work} summarizes this positioning.

\begin{table*}[t]
\centering
\small
\setlength{\tabcolsep}{2.5pt}
\caption{Positioning LegalCiteTrust against related paradigms. Citation trustworthiness requires legal authority existence, faithful description, and support for the current claim.}
\label{tab:related-work}
\begin{tabular}{p{0.18\textwidth}p{0.36\textwidth}p{0.06\textwidth}p{0.06\textwidth}p{0.06\textwidth}p{0.06\textwidth}p{0.06\textwidth}}
\toprule
Paradigm & Representative work & Legal & Deep \newline research & Long \newline report & Citation support & Citation trust \\
\midrule
Legal static & \begin{tabular}[t]{@{}l@{}}CAIL2018 \cite{xiao2018cail2018}\\LexGLUE \cite{chalkidis2022lexglue}\\LawBench \cite{fei2024lawbench}\end{tabular} & Yes & No & No & Limited & No \\
\addlinespace[4pt]
Legal retrieval & \begin{tabular}[t]{@{}l@{}}COLIEE \cite{goebel2023summary}\\LeCaRD \cite{ma2021lecard}\\CanLegalRAGBench \cite{zhao2026canlegalragbench}\end{tabular} & Yes & Partial & No & Partial & Partial \\
\addlinespace[4pt]
Legal agent & \begin{tabular}[t]{@{}l@{}}LegalAgentBench \cite{li2025legalagentbench}\\PLawBench \cite{shi2026plawbench}\end{tabular} & Yes & Partial & Partial & Partial & Partial \\
\addlinespace[4pt]
General agent & \begin{tabular}[t]{@{}l@{}}AgentBench \cite{liu2024agentbench}\\OpenHands Eval \cite{wang2025openhands}\end{tabular} & No & Partial & No & Limited & No \\
\addlinespace[4pt]
General deep research & \begin{tabular}[t]{@{}l@{}}ReportBench \cite{li2025reportbench}\\DeepResearch Bench \cite{bosse2025deep}\end{tabular} & No & Yes & Yes & Yes & No \\
\addlinespace[4pt]
Citation / hallucination & \begin{tabular}[t]{@{}l@{}}TruthfulQA \cite{lin2022truthfulqa}\\RAGAS \cite{es2024ragas}\\LegalCiteBench \cite{chen2026legalcitebench}\end{tabular} & Partial & Limited & Limited & Partial & Partial \\
\addlinespace[4pt]
This work & LegalCiteTrust & Yes & Yes & Yes & Yes & Yes \\
\bottomrule
\end{tabular}
\end{table*}

Legal benchmarks cover increasingly realistic legal capabilities. Static and understanding-oriented benchmarks such as CAIL2018~\cite{xiao2018cail2018}, LexGLUE~\cite{chalkidis2022lexglue}, and LawBench~\cite{fei2024lawbench} standardize evaluation for legal prediction, classification, and reasoning. Retrieval and grounding benchmarks such as COLIEE~\cite{goebel2023summary}, LeCaRD~\cite{ma2021lecard}, STARD~\cite{su2024stard}, and CanLegalRAGBench~\cite{zhao2026canlegalragbench} evaluate whether systems can locate or use legal sources. Practice-oriented and agentic legal benchmarks such as PLawBench~\cite{shi2026plawbench} and LegalAgentBench~\cite{li2025legalagentbench} move closer to realistic legal work products. These benchmarks are central to legal NLP evaluation, but they primarily assess legal competence, retrieval, answer correctness, or task completion rather than whether citations in a complete research report exist, are faithfully described, and support the local legal claim.

Agent and deep-research benchmarks evaluate broader multi-step research behavior. AgentBench~\cite{liu2024agentbench} and OpenHands-style evaluations~\cite{wang2025openhands} focus on planning, tool use, interaction, and task success. ReportBench~\cite{li2025reportbench}, DeepResearch Bench~\cite{bosse2025deep}, and DeepResearch Arena~\cite{wan2026deep} directly evaluate long-form research and report generation, including completeness, synthesis, factual consistency, and citation support. These settings are close to our report-level concern, but legal research reports impose a stronger notion of authority: statutes, judicial interpretations, regulations, and cases are not merely information sources, but normative grounds for legal argument.

Hallucination and citation-faithfulness evaluations address factuality and grounding from another direction. TruthfulQA~\cite{lin2022truthfulqa} and HaluEval~\cite{li2023halueval} evaluate factual or hallucinated content; RAGAS~\cite{es2024ragas}, CRAG~\cite{yang2024crag}, and RAGTruth~\cite{niu2024ragtruth} study faithfulness or hallucination in retrieval-grounded generation. Citation-focused work asks whether generated statements are supported by references, and LegalCiteBench~\cite{chen2026legalcitebench} targets legal citation generation and verification through citation recovery, matching, and correction. LegalCiteTrust targets the remaining intersection: long-form legal research reports in which citations must function as trustworthy legal authority, not merely as retrievable or matching references.

\section{Benchmark Design}
This section defines the task, benchmark construction, and citation-trustworthiness dimensions used by LegalCiteTrust. Appendix~\ref{app:metrics} provides the full metric mappings and aggregation rules.

\subsection{Task Definition}
LegalCiteTrust evaluates long-form legal research report generation. Given a task input consisting of a main legal research query, subtopics, and subtopic descriptions, a system is asked to produce a structured report that analyzes relevant legal rules, applicability conditions, cases, and normative authorities. The output is not a final answer alone: it is expected to organize issues across subtopics, explain legal grounds, and cite statutes, judicial interpretations, regulations, or cases as authority for the analysis.

A complete legal research report must do more than reach a plausible conclusion. It must cover the requested analysis, provide legal authority for that analysis, and use the cited authority reliably in context. LegalCiteTrust therefore maps these layers to Coverage, Support, and Citation Trustworthiness.

Coverage measures whether the report covers the rubric-level issues implied by the provided subtopics and descriptions. It is therefore an adherence measure under a given research plan, not open-ended legal issue spotting. Support measures evidence richness: whether the report supplies statutory, regulatory, judicial-interpretation, or case authority for relevant issues, without itself judging whether each citation is faithfully used. Citation Trustworthiness, quantified by a report-level Trust score, measures citation reliability through Existence, Fidelity, and Applicability (E/F/A) judgments. Separating these dimensions lets LegalCiteTrust distinguish whether a report covers the required analysis, provides legal evidence, and uses citations reliably.

\subsection{Dataset Construction}

We construct LegalCiteTrust from a candidate pool of 13,622 legal research documents. To ensure sufficient legal-authority demand for citation-trustworthiness evaluation, we retain documents with at least one case citation and at least five statutory or normative references, yielding 72 source documents.

\begin{table}[h]
\centering
\small
\caption{Dataset statistics.}
\label{tab:dataset-stats}
\begin{tabular}{lr}
\toprule
Statistic & Count \\
\midrule
Candidate source documents & 13,622 \\
Final source documents / tasks & 72 \\
Subtopics & 236 \\
Rubrics & 585 \\
Evidence points & 1,459 \\
\bottomrule
\end{tabular}
\end{table}

Each source document is converted into one legal research task with a main query, subtopics, and subtopic descriptions as generation input. Rubrics and evidence points are reserved for evaluation rather than shown to systems. Rubrics specify fine-grained discussion requirements, while evidence points provide reference material for judging whether a report covers and supports those requirements. The released benchmark materials consist of reconstructed task inputs, rubrics, evidence points, and annotations, and do not include the source-document full text, abstracts, or continuous original passages. Released materials were reviewed to avoid unnecessary personal identifying information; retained legal identifiers such as statute names, case names, docket numbers, or judgment references are included only when needed for citation verification. The resulting 72 tasks expand into 236 subtopics, 585 rubrics, and 1,459 evidence points, making the benchmark dense enough for controlled system comparison and ablation. Table~\ref{tab:dataset-stats} summarizes the benchmark structure.

\subsection{Citation Trustworthiness Dimensions}

Citation Trustworthiness is operationalized through three citation-level judgments: Existence, Fidelity, and Applicability. This decomposition separates fabricated or unresolved sources from more contextual legal failures: a report may cite a real authority while omitting a limiting condition, overstating the rule, or using a background source to support a stronger local claim.

Existence asks whether the cited legal source can be resolved, including law names, article numbers, normative documents, case names, docket numbers, or judgments. Fidelity asks whether the report faithfully describes the source content, including conditions, exceptions, facts, holdings, or legal rules. Applicability asks whether the source supports the local claim for which it is cited, rather than merely sharing the same topic or supporting a narrower proposition. Appendix~\ref{app:metrics} details the citation-verification workflow and aggregation rules, while Appendix~\ref{sec:qualitative-error-analysis} provides representative qualitative examples.

\begin{table*}[t]
\centering
\small
\setlength{\tabcolsep}{3pt}
\renewcommand{\arraystretch}{1.08}
\caption{Metric hierarchy and computation overview. Indentation indicates lower-level components. Score symbols are defined in Section~\ref{sec:evaluation-metrics}.}
\label{tab:metric-overview}
\begin{tabular}{p{0.25\textwidth}p{0.30\textwidth}p{0.34\textwidth}}
\toprule
Metric / component & Computation / aggregation & Meaning \\
\midrule
Final score & $(s_{\mathrm{cov}}+s_{\mathrm{sup}}) \times s_{\mathrm{trust}}$ & Completeness weighted by citation trustworthiness \\
\addlinespace[2pt]
\quad Coverage & Mean normalized rubric coverage & Adherence to the given research structure \\
\quad Support & $0.5 \times s_{\mathrm{rubric}} + 0.5 \times s_{\mathrm{subtopic}}$ & Legal evidence richness \\
\quad\quad Rubric Evidence Support & Mean normalized rubric support & Fine-grained evidence breadth \\
\quad\quad Subtopic Evidence Support & Mean normalized subtopic authority score & Subtopic-level law/case authority coverage \\
\quad Trust & Mean $E_iF_iA_i$ over included citation items & Reliability of included citations \\
\quad\quad Existence & Citation-level E score & Whether the cited source exists \\
\quad\quad Fidelity & Citation-level F score & Whether the source is faithfully described \\
\quad\quad Applicability & Citation-level A score & Whether the source supports the local claim \\
\bottomrule
\end{tabular}
\end{table*}

\begin{table*}[t]
\centering
\small
\setlength{\tabcolsep}{3pt}
\caption{Systems in the main evaluation.}
\label{tab:evaluated-systems}
\begin{tabular}{p{0.4\columnwidth}p{0.55\columnwidth}p{0.9\columnwidth}}
\toprule
Category & Systems & Role \\
\midrule
General LLM & \begin{tabular}[t]{@{}l@{}}GPT5~\cite{singh2025openai}\\DeepSeek~\cite{xu2026deepseek}\\Qwen36\_plus~\cite{yang2025qwen3}\end{tabular} & Direct report generation without additional search tools \\
Deep / agentic research & \begin{tabular}[t]{@{}l@{}}Deli\_DR\textsuperscript{a}\\Qwen\_DR~\cite{tongyidr}\end{tabular} & Multi-step research systems with denser evidence and citations \\
Legal-specific system & \begin{tabular}[t]{@{}l@{}}Deli\_LegalAI\textsuperscript{a}\\Farui\textsuperscript{b}\end{tabular} & End-to-end legal products or legal-domain systems \\
\bottomrule
\end{tabular}
\vspace{2pt}
\begin{minipage}{0.96\textwidth}
\footnotesize
\textsuperscript{a}Deli official product page: \url{https://www.delilegal.com/ai}.
\textsuperscript{b}Farui official product page: \url{https://tongyi.aliyun.com/farui/home}.
\end{minipage}
\end{table*}

\subsection{Evaluation Metrics}
\label{sec:evaluation-metrics}

LegalCiteTrust reports normalized report-level Coverage, Support, and Trust scores, denoted by $s_{\mathrm{cov}}$, $s_{\mathrm{sup}}$, and $s_{\mathrm{trust}}$. Table~\ref{tab:metric-overview} summarizes the metric hierarchy; Appendix~\ref{app:metrics} provides label mappings, normalization rules, and aggregation details.

Support combines rubric-level evidence support and subtopic-level authority support:
\begin{equation}
s_{\mathrm{sup}} = 0.5 \times s_{\mathrm{rubric}} + 0.5 \times s_{\mathrm{subtopic}}.
\end{equation}
Support is computed independently of E/F/A Trust scores: it measures evidence richness and authority coverage, while Trust measures citation reliability.

For each included law or case citation item $i$, Existence, Fidelity, and Applicability scores are multiplied and averaged into report-level Trust:
\begin{equation}
s_{\mathrm{trust}} =
\frac{1}{N_{\mathrm{cite}}}
\sum_{i=1}^{N_{\mathrm{cite}}} E_iF_iA_i .
\end{equation}
If a report contains no included law/case citation items, $s_{\mathrm{trust}}$ is set to 1.0 because Trust is conditional on included citations; citation scarcity is captured by Support and citation-count diagnostics.

The final report score combines report completeness and citation trustworthiness:
\begin{equation}
s_{\mathrm{final}} = (s_{\mathrm{cov}} + s_{\mathrm{sup}}) \times s_{\mathrm{trust}}.
\end{equation}
Since Coverage, Support, and Trust are normalized to $[0,1]$, $s_{\mathrm{final}}$ ranges from 0 to 2 and is not further normalized.

Main leaderboard scores use report-level macro averages, while item-level E/F/A diagnostic tables use citation-level micro averages. Thus, leaderboard tables compare task-level system performance, whereas item-level diagnostics describe the average reliability of generated law and case references.

\begin{table*}[t]
\centering
\scriptsize
\setlength{\tabcolsep}{3.5pt}
\caption{Main leaderboard report-level summary. Report-level E/F/A averages summarize citation verification within reports.}
\label{tab:main-report-summary}
\begin{tabular}{l r r r r r r r r r}
\toprule
\multirow{2}{*}{Model} & \multirow{2}{*}{Final} & \multicolumn{3}{c}{Main dimensions} & \multicolumn{2}{c}{Support components} & \multicolumn{3}{c}{Report-level E/F/A} \\
\cmidrule(lr){3-5}\cmidrule(lr){6-7}\cmidrule(lr){8-10}
& & Cov & Sup & Trust & Rubric & Subtopic & E & F & A \\
\midrule
Deli\_DR & 0.9818 & 0.9204 & 0.3003 & 0.8064 & 0.3390 & 0.2616 & 0.9270 & 0.9487 & 0.9086 \\
GPT5 & 0.9776 & 0.9525 & 0.1895 & 0.8575 & 0.3111 & 0.0679 & 0.8600 & 0.9451 & 0.9091 \\
Qwen36\_plus & 0.9612 & 0.9248 & 0.3153 & 0.7744 & 0.3164 & 0.3142 & 0.9011 & 0.9489 & 0.9079 \\
DeepSeek & 0.9591 & 0.9367 & 0.2604 & 0.8077 & 0.2684 & 0.2524 & 0.9168 & 0.9452 & 0.9039 \\
Deli\_LegalAI & 0.9221 & 0.8727 & 0.2736 & 0.8041 & 0.2944 & 0.2528 & 0.9343 & 0.9313 & 0.9182 \\
Qwen\_DR & 0.8856 & 0.8424 & 0.4020 & 0.7164 & 0.3795 & 0.4245 & 0.8630 & 0.9212 & 0.8867 \\
Farui & 0.3835 & 0.3963 & 0.0300 & 0.9133 & 0.0215 & 0.0385 & 0.9383 & 0.9580 & 0.9084 \\
\bottomrule
\end{tabular}
\end{table*}

\section{Experimental Setup}
We evaluate LegalCiteTrust in three settings: a main leaderboard over representative legal research systems, a tool ablation that varies external search access, and a verification ablation that tests whether citation-verification feedback can improve generated reports.

The main leaderboard evaluates seven end-to-end systems spanning general-purpose LLMs, deep or agentic research systems, and legal-specific systems. We compare realistic system paradigms rather than isolating base-model capability, because retrieval access, legal specialization, and workflow design are part of deployed legal research behavior. Table~\ref{tab:evaluated-systems} summarizes the evaluated systems.

All systems are evaluated on the same 72 tasks. For each task, the system receives the same main query, subtopics, and subtopic descriptions, and is asked to generate a complete legal research report. Retrieval conditions follow each system's realistic usage mode: direct LLMs receive no additional tools, deep-research systems use their own research workflows, and legal-specific systems use their built-in legal retrieval or research assistance. For API-based systems, we use default model parameters. All outputs are evaluated by the LegalCiteTrust pipeline, producing Coverage, Support, Trust, and Final scores; the tool and verification ablation details are introduced with their results in Sections~5.2 and~5.3. Additional computational and implementation details are provided in Appendix~\ref{app:computational-implementation-details}.

\section{Results}
The main text reports report-level summaries for the main leaderboard, tool ablation, and verification ablation. These tables compare task-level system performance using Final, Coverage, Support, Trust, and report-level E/F/A averages. Full item-level E/F/A diagnostics, computed as citation-level micro averages over generated law and case references, are provided in Appendix~\ref{app:item-diagnostics}.

\begin{table*}[t]
\centering
\scriptsize
\setlength{\tabcolsep}{3.5pt}
\caption{Tool ablation report-level summary. Report-level E/F/A averages summarize citation verification within reports.}
\label{tab:tool-report-summary}
\begin{tabular}{l r r r r r r r r r}
\toprule
\multirow{2}{*}{Model} & \multirow{2}{*}{Final} & \multicolumn{3}{c}{Main dimensions} & \multicolumn{2}{c}{Support components} & \multicolumn{3}{c}{Report-level E/F/A} \\
\cmidrule(lr){3-5}\cmidrule(lr){6-7}\cmidrule(lr){8-10}
& & Cov & Sup & Trust & Rubric & Subtopic & E & F & A \\
\midrule
Qwen36\_plus & 0.9612 & 0.9248 & 0.3153 & 0.7744 & 0.3164 & 0.3142 & 0.9011 & 0.9489 & 0.9079 \\
Qwen36\_tool\_law & 0.9718 & 0.8855 & 0.3427 & 0.7951 & 0.3344 & 0.3510 & 0.9289 & 0.9495 & 0.8917 \\
Qwen36\_tool\_law\_case & 1.0329 & 0.9065 & 0.4857 & 0.7433 & 0.4568 & 0.5146 & 0.9106 & 0.9198 & 0.8752 \\
Qwen36\_tool\_law\_case\_web & 0.9903 & 0.9074 & 0.4197 & 0.7441 & 0.4128 & 0.4266 & 0.8965 & 0.9438 & 0.8824 \\
\bottomrule
\end{tabular}
\end{table*}

\begin{table*}[t]
\centering
\scriptsize
\setlength{\tabcolsep}{3.5pt}
\caption{Verification ablation report-level summary. Report-level E/F/A averages summarize citation verification within reports.}
\label{tab:verification-report-summary}
\begin{tabular}{l r r r r r r r r r}
\toprule
\multirow{2}{*}{Model} & \multirow{2}{*}{Final} & \multicolumn{3}{c}{Main dimensions} & \multicolumn{2}{c}{Support components} & \multicolumn{3}{c}{Report-level E/F/A} \\
\cmidrule(lr){3-5}\cmidrule(lr){6-7}\cmidrule(lr){8-10}
& & Cov & Sup & Trust & Rubric & Subtopic & E & F & A \\
\midrule
Base & 1.0329 & 0.9065 & 0.4857 & 0.7433 & 0.4568 & 0.5146 & 0.9106 & 0.9198 & 0.8752 \\
E-Filter & 1.0970 & 0.9099 & 0.4487 & 0.8134 & 0.4415 & 0.4559 & 0.9980 & 0.9160 & 0.8673 \\
EFA-Revise & 1.1708 & 0.9011 & 0.4223 & 0.8883 & 0.4177 & 0.4270 & 0.9836 & 0.9713 & 0.9285 \\
\bottomrule
\end{tabular}
\end{table*}

\subsection{Main Leaderboard}
Table~\ref{tab:main-report-summary} reports the main leaderboard. No system dominates all dimensions: Final, Coverage, Support, Trust, and report-level E/F/A averages reveal different system behaviors. Deli\_DR obtains the highest Final score by maintaining a balanced profile across Coverage, Support, and Trust, while GPT5 achieves high Coverage and Trust but much lower Support, especially at the subtopic level. Qwen\_DR shows the opposite pattern: it has the highest Support and densest citation behavior, but substantially lower Trust.

Farui illustrates why Trust alone is insufficient. It achieves the highest Trust, but its Coverage and Support are both very low, leading to a much lower Final score. Conversely, Qwen\_DR shows that high evidence support does not guarantee citation reliability. Item-level diagnostics in Appendix~\ref{app:item-diagnostics} further show that Qwen\_DR generates the largest number of both law and case citations, explaining its high Support while also increasing exposure to citation errors. The main leaderboard therefore motivates evaluating long-form legal research systems jointly along Coverage, Support, and Trust.

The item-level view also shows that most systems generate many more law citations than case citations, so case-level E/F/A scores should be interpreted cautiously when $n_{\mathrm{case}}$ is small. This imbalance is itself informative: legal research systems can appear well grounded through abundant statutory references while still struggling to use cases consistently as applicable authority.

\subsection{Tool Ablation}
The tool ablation isolates how retrieval access changes one base system. Starting from Qwen36\_plus, we progressively add law search, law plus case search, and law plus case plus web search, while keeping the task input and base prompt fixed. Table~\ref{tab:tool-report-summary} reports the report-level summary.

Law search modestly increases Support from 0.3153 to 0.3427 and Trust from 0.7744 to 0.7951, raising Final from 0.9612 to 0.9718. Adding case search produces the largest evidence gain: Support rises to 0.4857, especially through Subtopic Evidence Support, but Trust drops to 0.7433. The law-plus-case condition still achieves the highest Final score among tool settings, showing that case evidence improves the composite score while increasing citation risk.

Adding web search does not further improve performance: Support falls to 0.4197, Trust remains nearly unchanged, and Final drops to 0.9903. Item-level diagnostics in Appendix~\ref{app:item-diagnostics} clarify this pattern: case citations increase from 4 in the base setting to 146 with case search, then fall to 73 when web search is added. Tool-call statistics point in the same direction: the law-only setting uses about 1.97 tool calls per report on average; law plus case uses about 4.15, including about 2.24 case-search calls; and law plus case plus web uses about 6.10, including about 1.94 web-search calls. Tool augmentation therefore shifts the balance between evidence richness and citation reliability rather than producing monotonic improvement.

\subsection{Verification Ablation}

The verification ablation tests whether E/F/A signals can improve generated reports rather than merely score them after the fact. We use Qwen36\_tool\_law\_case as the Base system, add an LLM revision stage, and compare two feedback variants: E-Filter receives only Existence results, while EFA-Revise receives the full E/F/A verification report. All three settings use the same 72 tasks, and Table~\ref{tab:verification-report-summary} reports the report-level summary.

Trust increases monotonically from Base to E-Filter and EFA-Revise: 0.7433, 0.8134, and 0.8883. Final also rises from 1.0329 to 1.0970 and 1.1708, while Coverage remains broadly stable and Support decreases. This indicates that verification feedback improves citation reliability by removing or revising unreliable evidence, even when this sacrifices some evidence support.

The comparison also shows why full E/F/A feedback is more useful than existence-only filtering. E-Filter mainly removes nonexistent or unresolved citations; Appendix~\ref{app:item-diagnostics} shows that law E rises from 0.9446 to 0.9987 and case E rises from 0.8014 to 0.9808, but case F and case A decrease after filtering. EFA-Revise uses richer feedback about source description and claim support, improving law F, law A, case F, and case A relative to Base and producing the largest Trust gain. Representative cases in Appendix~\ref{sec:qualitative-error-analysis} illustrate how Existence, Fidelity, and Applicability errors differ in generated reports. E/F/A therefore functions not only as a benchmark diagnostic, but also as a feedback signal for verification-aware legal research systems.

\section{Discussion}
The analyses in this section are descriptive system-level summaries rather than formal significance tests. The plotted points combine main-leaderboard systems with non-duplicate tool-ablation settings, and exclude verification-ablation settings because those are post-hoc revision variants analyzed separately in Section~5.3. Farui remains in the leaderboard as a deployed legal product, but is excluded from the main scatter plots because its sparse, low-Coverage outputs form a boundary case rather than a comparable full-report behavior. Appendix~\ref{app:trend-robustness} reports the corresponding robustness checks.

\subsection{Coverage-Like Evaluation Does Not Capture Citation Trustworthiness}
\begin{figure*}[t]
\centering
\begin{minipage}{0.48\textwidth}
\centering
\includegraphics[width=\linewidth]{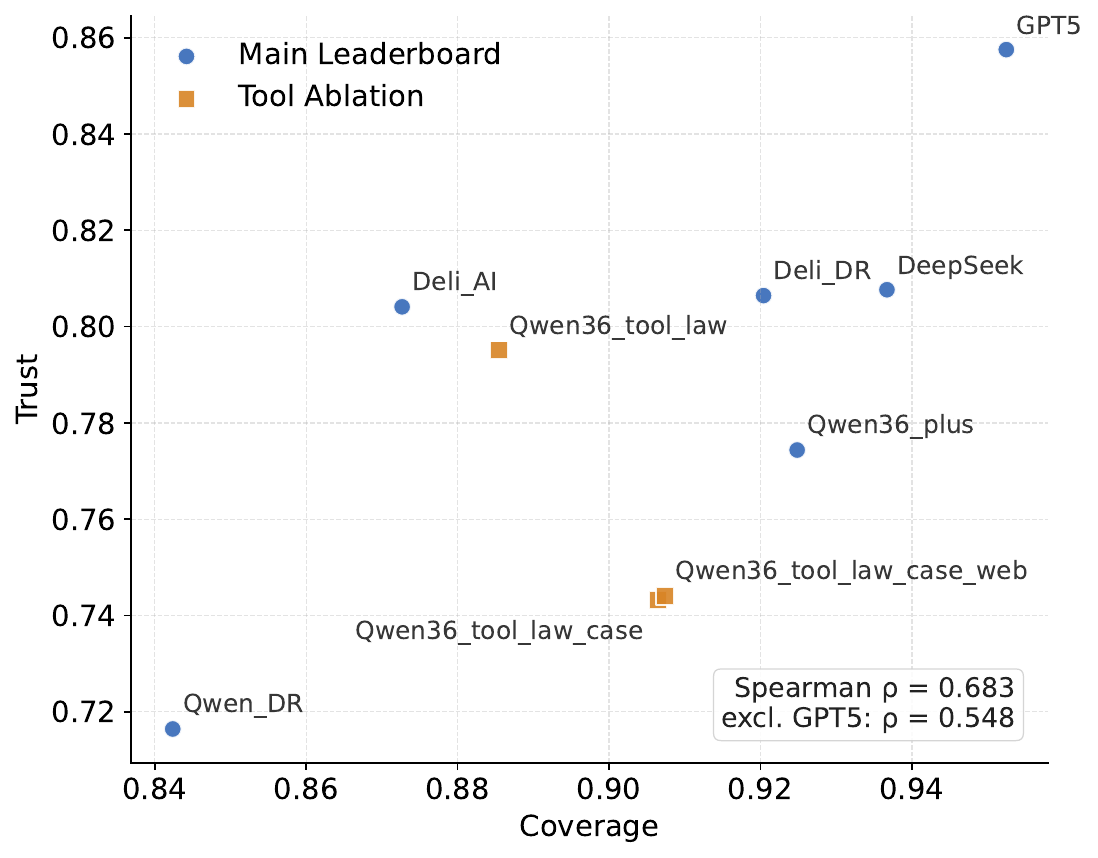}
\end{minipage}
\hfill
\begin{minipage}{0.48\textwidth}
\centering
\includegraphics[width=\linewidth]{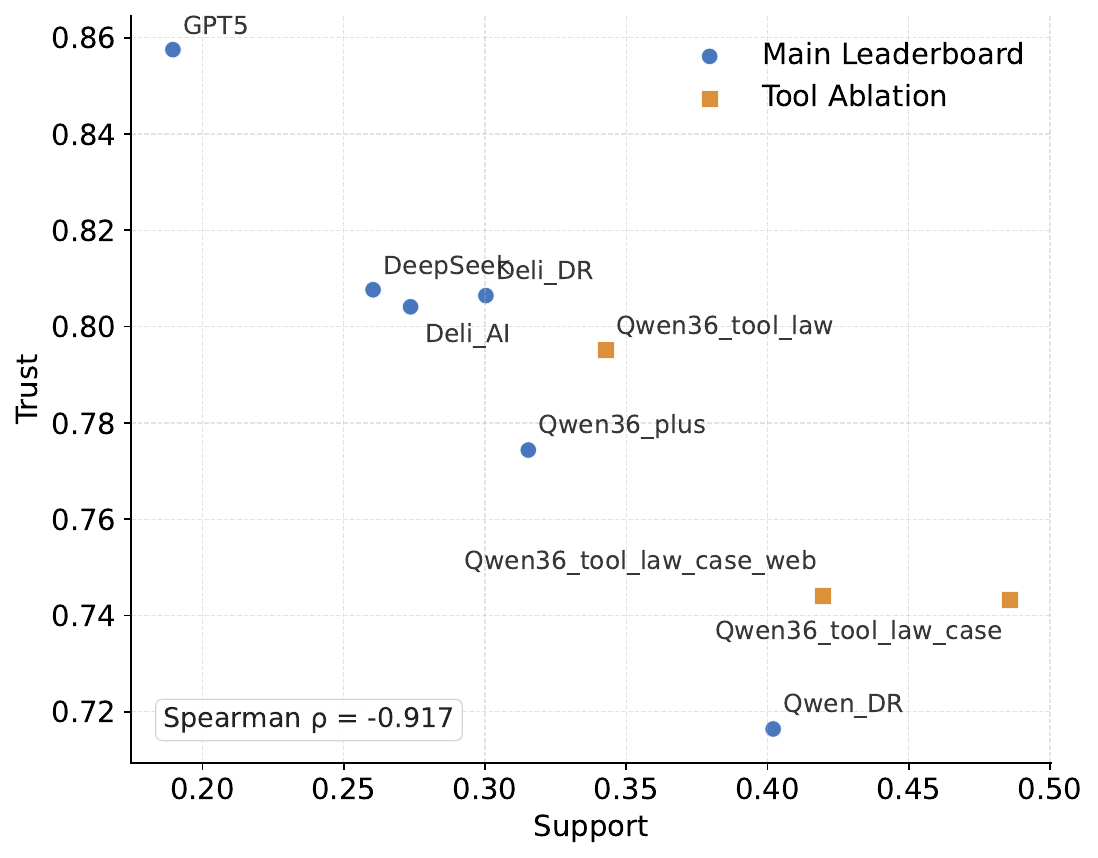}
\end{minipage}
\caption{Coverage-like report completion and evidence support capture different aspects of citation trustworthiness. Left: Coverage is only moderately associated with Trust. Right: richer evidence support is negatively associated with Trust, indicating that authority coverage and citation reliability can pull in different directions.}
\label{fig:coverage-support-trust}
\end{figure*}

Many long-form and practice-oriented benchmarks emphasize task completion, topical coverage, answer usefulness, or overall report quality \cite{li2025reportbench,bosse2025deep}. These dimensions are important, but they are closer to whether a report covers the requested research structure than to whether its legal authorities are trustworthy. LegalCiteTrust makes this distinction explicit by separating Coverage from Trust. In the left panel of Figure~\ref{fig:coverage-support-trust}, Coverage and Trust are only moderately associated ($\rho=0.6833$; excluding GPT5: $\rho=0.5476$). Thus, coverage-like report evaluation is necessary but not sufficient: a report can cover the requested issues while still misusing legal authorities, and a high-Trust output can be incomplete if it avoids difficult issues or cites very little.

The right panel of Figure~\ref{fig:coverage-support-trust} shows the complementary Support--Trust tension. Systems that provide richer legal evidence tend to incur lower citation trustworthiness ($\rho=-0.9167$). This does not mean that evidence support is undesirable; rather, when systems produce more specific statutes, cases, and legal authorities, they also create more opportunities for Existence, Fidelity, and Applicability errors.

\subsection{More Evidence Requires Citation-Aware Governance}
The evidence-governance interpretation is supported by three observations. First, the tool ablation shows that search tools can increase evidence support, especially when case search is available, but do not automatically improve Trust. Second, the negative Support--Trust trend is present in both rubric-level and subtopic-level support components. Third, the verification ablation shows that E/F/A signals can improve reports rather than merely score them: Trust rises from 0.7433 for the Base system to 0.8134 with existence-only filtering and to 0.8883 with full E/F/A revision, while Final increases from 1.0329 to 1.0970 and 1.1708.

\begin{figure}[t]
\centering
\includegraphics[width=\columnwidth]{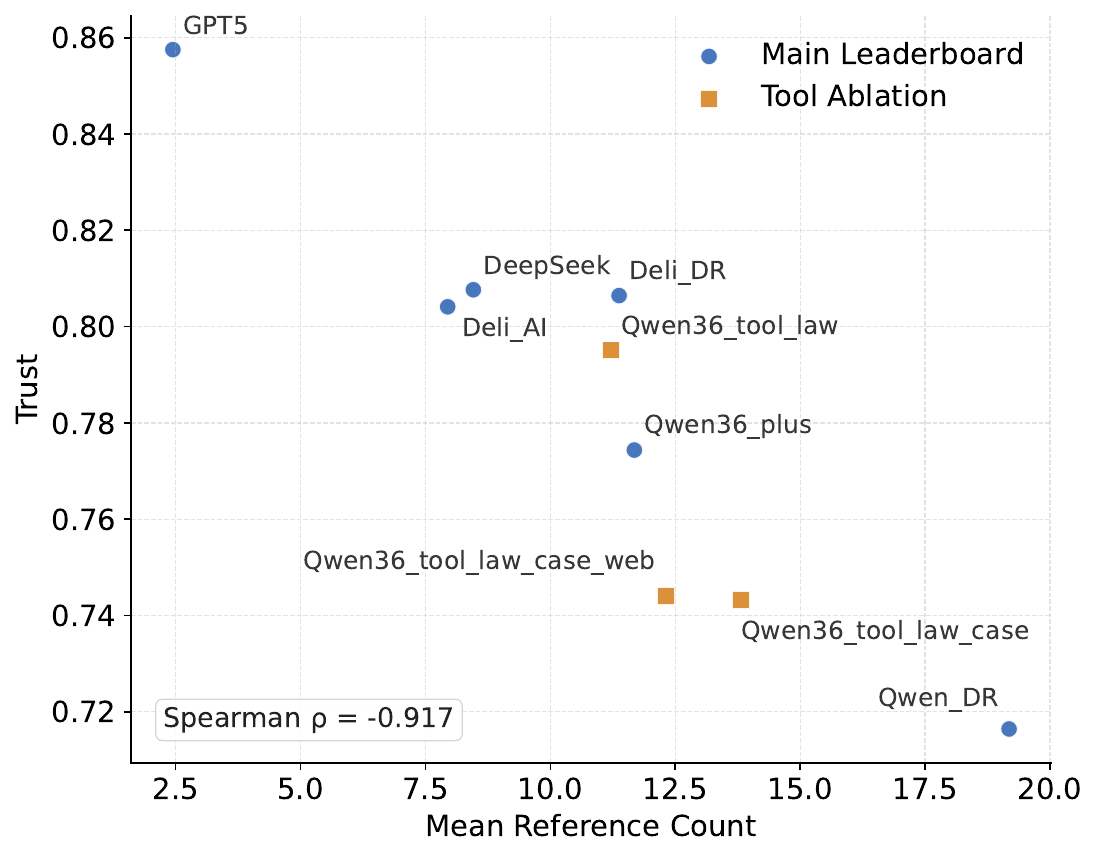}
\caption{System-level relationship between reference count and Trust. Denser citation behavior increases verification burden.}
\label{fig:reference-count-trust}
\end{figure}

Citation density provides a concrete mechanism behind this tension. Figure~\ref{fig:reference-count-trust} shows a strong negative association between reference count and Trust ($\rho=-0.9167$). This should not be read as ``fewer citations are better'': legal research reports need sufficient authority. The point is that every added statute, regulation, or case must satisfy Existence, Fidelity, and Applicability. If a system lacks reliable retrieval constraints, source selection, and semantic verification, denser citation behavior can amplify both the number of errors and their visibility.

These results indicate that post-retrieval reliability depends on more than acquiring additional authorities. Retrieval and agentic search expand the pool of candidate statutes, regulations, and cases; trustworthy legal research generation also requires mechanisms that govern which authorities are selected, how they are described, and whether they support the local claims for which they are cited. The central challenge is therefore not retrieval augmentation alone, but citation-aware evidence governance after retrieval. Evaluation reliability checks support this interpretation: human audit, repeated-judge checks, and bootstrap resampling show that the main E/F/A and leaderboard trends are stable; Appendix~\ref{app:evaluation-reliability} reports the full protocol and results.

\section{Conclusion}
We introduced LegalCiteTrust, a benchmark for citation trustworthiness in Chinese long-form legal research reports, operationalized through Existence, Fidelity, and Applicability and evaluated together with Coverage and Support. Across 72 tasks and multiple system types, we find that report completion, evidence richness, citation density, and citation reliability expose different behaviors: systems can cover the requested structure or cite more authorities while still misusing legal sources.

The verification ablation shows that E/F/A signals are useful not only for evaluation but also for revision, improving Trust and Final more clearly than existence-only filtering. These results point to citation-aware evidence governance as a central requirement for trustworthy legal research generation: systems must not only retrieve legal authorities, but also select, describe, and apply them reliably. Future work can adapt this framework across jurisdictions and integrate citation verification more tightly into generation.

\section{Limitations}
LegalCiteTrust turns citation trustworthiness into an explicit evaluation protocol, but it does not make legal citation use mechanically objective. Existence is largely a source-resolution question, whereas Fidelity and Applicability are protocolized legal judgments about conditions, exceptions, factual context, statute versions, and the distance between a cited source and the local claim. Boundary cases may arise when a report omits a minor qualification, underspecifies the applicable source version, or cites a source that supports a narrower proposition but not the stronger claim made in the report. Human audit, repeated-judge checks, and bootstrap analyses test consistency with the protocol, not agreement with objective legal truth. LegalCiteTrust should therefore support expert review rather than replace it; Appendix~\ref{app:societal-risks} discusses the corresponding overuse risk in legal settings.

The benchmark is dense rather than broad: its 72 tasks are authority-demanding Chinese legal research problems that expand into many subtopics, rubrics, evidence points, and verification items. This design supports controlled comparison and ablation, but it does not cover all legal topics, report genres, or user intents. Transfer to other jurisdictions is also nontrivial because citation formats, statute versioning, case authority, and legal research practice vary across legal systems. Common-law settings, for example, may require different Applicability criteria for precedential force, factual analogy, and treatment history.

Verification is bounded by independently checkable legal sources and by the coverage of reference stores and commercial retrieval APIs. A valid citation missing from these sources may be treated as unverifiable, which can disadvantage closed systems but keeps citation checking reproducible. This choice reflects a core requirement for citation trustworthiness: readers and downstream users should be able to independently check the legal authorities cited in a report.

Finally, LegalCiteTrust evaluates citation trustworthiness rather than the full quality of legal advice. A report with accurate and applicable citations may still be weak in legal strategy, risk assessment, persuasive structure, readability, or adaptation to a client's objectives. The main leaderboard also compares realistic end-to-end research systems rather than identical base models. Retrieval infrastructure, product workflows, prompting, and post-processing can all affect performance, so the tool and verification ablations should be read as mechanism probes rather than a complete causal decomposition.

\bibliography{references}

\appendix

\section{Metric Details}
\label{app:metrics}
This appendix provides the metric definitions and scoring details behind Section~3. The main paper keeps only the core formulas and design motivation for Coverage, Support, Trust, and Final. Here we specify how rubric-level, subtopic-level, and citation-level evaluation units are aggregated into report-level and system-level scores.

The three main dimensions are computed independently. Coverage measures whether a report covers the rubric-level discussion requirements in the given research structure. Support measures whether the report provides legal evidence for rubrics and subtopics. Trust measures whether included law and case citation items satisfy Existence, Fidelity, and Applicability. The final report score is
\[
\mathrm{Final}=(\mathrm{Coverage}+\mathrm{Support})\times \mathrm{Trust}.
\]
Coverage, Support, and Trust are normalized to $[0,1]$. Final is a raw composite score with theoretical range $[0,2]$ and is not additionally normalized.

\subsection{Overview}
Each report is scored through five steps. First, the evaluator assigns a 0/1/2/3 Coverage score to each rubric and averages these scores within the report after normalization. Second, the evaluator separately assigns law-support and case-support scores for each rubric; Rubric Evidence Support takes the larger of the two scores and then averages over rubrics. Third, law and case authorities in the report are assigned to subtopics; Subtopic Evidence Support is computed from the number and type of countable authorities under each subtopic, rather than from E/F/A label scores. Fourth, each included law or case citation item receives Existence, Fidelity, and Applicability scores, which are multiplied into a reference-level Trust score and pooled into report-level Trust. Fifth, Coverage, Support, and Trust are combined into Final.

System-level leaderboard scores are report-level macro averages, giving each report equal weight. When report-level summary tables include E/F/A columns, the E/F/A layer scores are first aggregated within each report and then macro-averaged over reports. The dedicated item-level E/F/A diagnostic tables instead use citation-level micro averages, giving each included citation item equal weight. Thus the leaderboard compares task-level system performance, while the item-level diagnostic tables describe the average reliability of concrete law and case references generated by a system.

\subsection{Coverage}
Coverage measures whether a report covers the rubric-level discussion requirements for a task. Because the generation input already contains subtopics and subtopic descriptions, Coverage should be interpreted as adherence to the given research structure rather than open-ended legal issue spotting.

\begin{table}[t]
\centering
\small
\caption{Rubric-level Coverage scoring.}
\label{tab:app-coverage-score}
\begin{tabular}{cp{0.76\columnwidth}}
\toprule
Score & Definition \\
\midrule
0 & Does not cover the core content of the rubric. \\
1 & Mentions only keywords, concepts, or conclusions. \\
2 & Clearly explains the meaning, cause, manifestation, or applicable scenario of the research point. \\
3 & Provides paragraph-level analysis with at least two types of elaboration, such as causes, consequences, judicial application, or typological criteria. \\
\bottomrule
\end{tabular}
\end{table}

For a report with $N_{\mathrm{rubric}}$ rubrics, let $\mathrm{coverage}_i$ be the score of rubric $i$. Report-level Coverage is
\[
\mathrm{Coverage}=\frac{\sum_i \mathrm{coverage}_i}{3N_{\mathrm{rubric}}}.
\]
This normalization places Coverage in $[0,1]$.

\subsection{Support}
Support measures whether a report provides legal evidence for the given research structure. It combines Rubric Evidence Support (RES) and Subtopic Evidence Support (SES):
\[
\mathrm{Support}=0.5\times \mathrm{RES}+0.5\times \mathrm{SES}.
\]
The two components differ in both evaluation level and scoring mechanism. Rubric Evidence Support is rubric-level evidence coverage judged for each rubric. Subtopic Evidence Support is subtopic-level authority coverage computed by assigning law and case authorities to subtopics and then applying count-based post-processing rules. Both components are independent of E/F/A Trust scores.

\subsubsection{Rubric Evidence Support}
Rubric Evidence Support measures whether each rubric receives legal-evidence support. It does not verify whether the cited statute or case exists, is faithfully described, or supports the local claim; these citation-reliability questions are handled by Trust.

For each rubric $i$, let $l_i$ and $c_i$ denote the law-support and case-support scores. The rubric support score is
\[
r_i=\max(l_i,c_i).
\]
The label meanings are summarized in Table~\ref{tab:app-rubric-support}.

\begin{table}[t]
\centering
\small
\caption{Rubric-level evidence-support scoring.}
\label{tab:app-rubric-support}
\begin{tabular}{cp{0.75\columnwidth}}
\toprule
Score & Definition \\
\midrule
\multicolumn{2}{l}{\textit{Law support}} \\
0 & No statute, judicial interpretation, or normative document supports the rubric. \\
1 & A law name or article number is present, but only as background or with weak connection. \\
2 & A law/document name and article number are clearly used for the rubric's core judgment. \\
3 & In addition to score 2, the report paraphrases the rule or explains validity, application path, or adjudicative effect. \\
\midrule
\multicolumn{2}{l}{\textit{Case support}} \\
0 & No case authority supports the rubric. \\
2 & An identifiable case is clearly used for the rubric's core judgment. \\
3 & An identifiable case is used and the report also states case facts, judicial reasoning, conclusion, or rule. \\
\bottomrule
\end{tabular}
\end{table}

For a report with $N_{\mathrm{rubric}}$ rubrics, report-level Rubric Evidence Support is
\[
\mathrm{RES}=\frac{\sum_i r_i}{3N_{\mathrm{rubric}}}.
\]

\subsubsection{Subtopic Evidence Support}
Subtopic Evidence Support measures whether each subtopic has sufficient law/case authorities. It does not directly use E/F/A label scores and does not require an authority's Fidelity or Applicability score to pass a threshold. Its role is to capture authority coverage at the subtopic level, while Trust separately measures whether those citations are reliable.

The computation has two steps. First, law and case authorities in a report are assigned to subtopics or to an ``other'' category. Second, post-processing code computes a raw score from the number and type of countable authorities under each subtopic. An authority is countable if it has extractable factual/citation context or claim-support content. Thus, an authority can contribute to Subtopic Evidence Support even if a later E/F/A judge assigns a partial or failing F/A score; citation reliability is handled separately by Trust.

\begin{table}[t]
\centering
\small
\caption{Subtopic Evidence Support raw scoring.}
\label{tab:app-subtopic-support}
\begin{tabular}{@{}cp{0.64\columnwidth}@{}}
\toprule
Raw score & Definition \\
\midrule
0 & No countable authority. \\
1 & One countable authority. \\
2 & Two or more countable authorities. \\
3 & Three or more countable authorities, including both law and case authorities. \\
\bottomrule
\end{tabular}
\end{table}

For each subtopic $j$, let $a_j$ be the raw score. The normalized score is
\[
u_j=\frac{a_j}{3}.
\]
For a report with $N_{\mathrm{subtopic}}$ subtopics,
\[
\mathrm{SES}=\frac{\sum_j u_j}{N_{\mathrm{subtopic}}}.
\]

\subsubsection{Overall Support}
Overall Support equally combines rubric-level evidence breadth and subtopic-level authority coverage. This design captures both fine-grained evidence support and higher-level law/case authority coverage, while leaving citation reliability errors to Trust.

\subsection{Trust Score}
Trust measures the reliability of included law and case citation items in a report. It is not a score for overall legal-opinion quality or citation quantity. Instead, it asks whether the citations already used by the report exist, are faithfully described, and are applicable to the local claims they support.

\subsubsection{Citation Verification Workflow}
\begin{figure*}[t]
\centering
\includegraphics[width=0.90\textwidth]{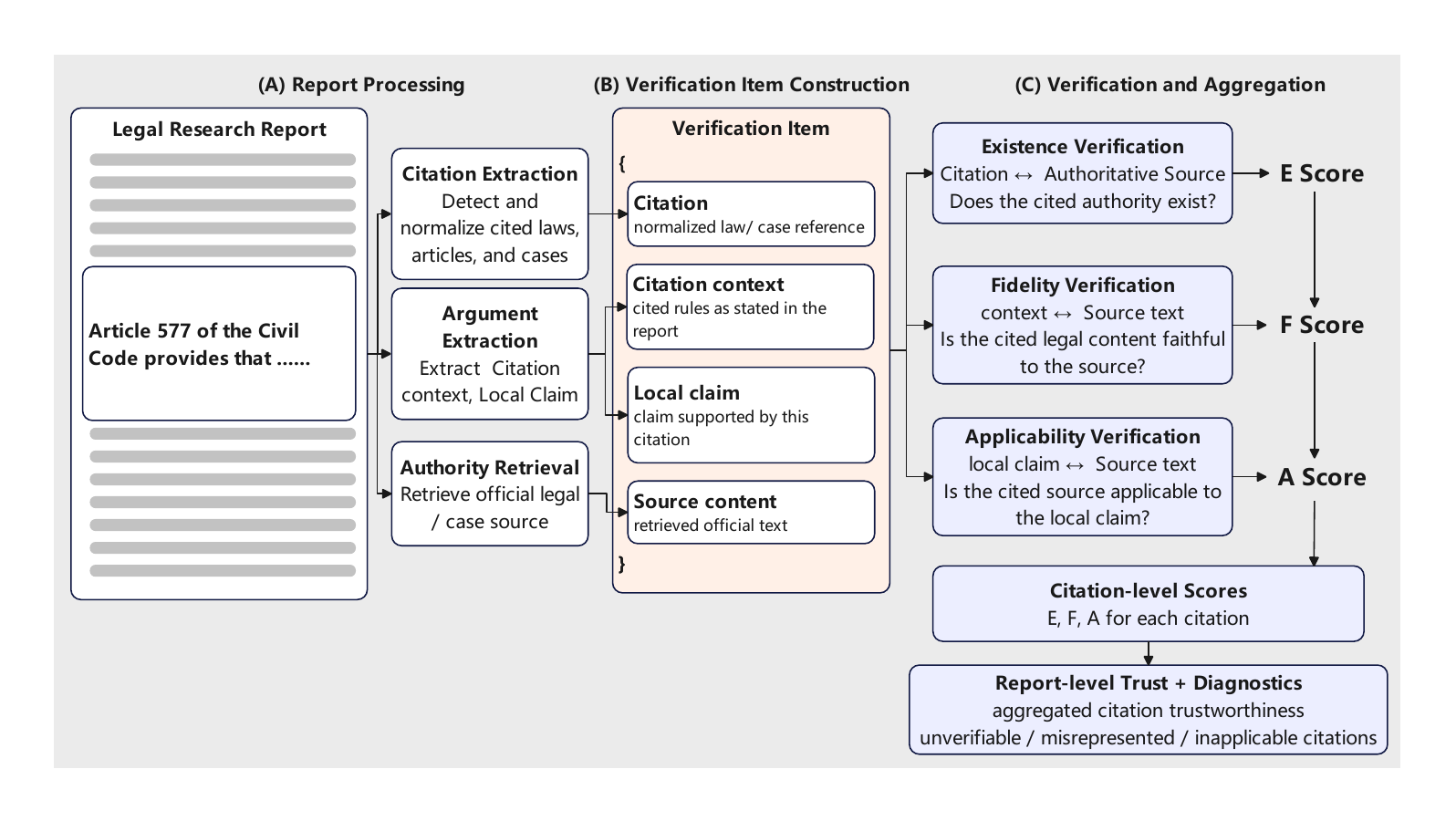}
\caption{High-level citation verification protocol in LegalCiteTrust. Model-generated reports are converted into citation-level verification items and evaluated through Existence, Fidelity, and Applicability before aggregation into report-level Trust score and diagnostics.}
\label{fig:verification-pipeline}
\end{figure*}

Given a generated report, LegalCiteTrust first identifies legal references and normalizes them into law and case citations. Each citation is converted into a verification item containing the cited reference, its report context, the local claim it is used to support, and candidate source content retrieved from legal sources. Local claims and citation content may be extracted with LLM assistance, but extraction is constrained to the report context and retrievable legal sources. Figure~\ref{fig:verification-pipeline} summarizes this flow.

The protocol applies E/F/A judgments sequentially. Existence checks whether the cited object can be resolved in legal source collections or public legal retrieval services. Fidelity checks whether the report's description faithfully reflects the source content. Applicability checks whether the source supports the local claim rather than merely sharing the same topic. Law and case citations follow the same E/F/A framework but use separate item structures because their identifiers, source contents, and support logic differ. Appendix~\ref{app:evaluation-reliability} reports reliability checks for the automatic evaluation.

\subsubsection{Citation Item Construction and Inclusion}
Each law or case citation is converted into a citation-level verification item. A verification item typically includes the citation text or normalized reference, the citation context in the report, the local claim supported by the citation, and retrievable source content.

Not every extracted reference enters the Trust denominator. Some references are marked as excluded, such as non-target laws, foreign laws, draft or consultation documents, or cases excluded during case resolution. Excluded items do not contribute to item-level Trust or to E/F/A average-score denominators.

\subsubsection{General E/F/A Formula}
For each included citation item, the reference-level Trust score is
\[
\mathrm{item\_trust}=E\times F_{\mathrm{NA}=1}\times A_{\mathrm{NA}=1}.
\]
Here $E$ is the Existence score; $F_{\mathrm{NA}=1}$ is the Fidelity score, with NA or skipped Fidelity treated as 1.0; and $A_{\mathrm{NA}=1}$ is the Applicability score, with NA or skipped Applicability treated as 1.0. This means that an unavailable F/A judgment does not add an extra penalty, while Existence must always be scored. Unknown labels not covered by the score mapping are treated as 0.0.

\subsubsection{Law Citation Scoring}
Law citations include statutes, judicial interpretations, and normative documents. Law E is determined from law resolution, article resolution, and legal status. A law citation receives full Existence credit only when both the law and article are resolved and the law is effective. Expired, pending, or timeless status receives partial credit; hallucinated, unresolved, or nonexistent article results receive zero. Non-target laws, foreign laws, drafts, and consultation documents are excluded from law Trust scoring.

\begin{table}[t]
\centering
\small
\caption{Law Existence scoring.}
\label{tab:app-law-e}
\begin{tabular}{p{0.72\columnwidth}c}
\toprule
Condition & Score \\
\midrule
Law resolved, article resolved, status effective & 1.0 \\
Law resolved, article resolved, status expired, pending, or no timeliness & 0.5 \\
Suspected hallucinated law or undetermined law conclusion & 0.0 \\
Article conclusion indicates nonexistent article & 0.0 \\
Law or article not resolved & 0.0 \\
\bottomrule
\end{tabular}
\end{table}

\begin{table}[t]
\centering
\small
\caption{Law Fidelity and Applicability label-to-score mapping.}
\label{tab:app-law-fa}
\begin{tabular}{p{0.54\columnwidth}c}
\toprule
Label & Score \\
\midrule
\multicolumn{2}{l}{\textit{Law Fidelity}} \\
Consistent & 1.0 \\
Minor omission & 1.0 \\
Key omission & 0.5 \\
Paraphrase deviation & 0.5 \\
Clear conflict & 0.0 \\
Irrelevant & 0.0 \\
Insufficient input & NA, as 1.0 \\
Other unknown label & 0.0 \\
\midrule
\multicolumn{2}{l}{\textit{Law Applicability}} \\
Supports & 1.0 \\
Partially supports & 0.5 \\
Overextended inference & 0.5 \\
Contradicts & 0.0 \\
Irrelevant & 0.0 \\
Insufficient input & NA, as 1.0 \\
Other unknown label & 0.0 \\
\bottomrule
\end{tabular}
\end{table}

Law F evaluates whether the report faithfully describes the statute or normative source content. Law A evaluates whether the source applies to and supports the report's local claim.

\subsubsection{Case Citation Scoring}
Case citations include case names, docket numbers, judgments, or other identifiable case information. Case E is binary in the current implementation: matched cases receive 1.0, excluded cases are removed from case Trust scoring, and unresolved, unmatched, empty, or other statuses receive 0.0.

\begin{table}[t]
\centering
\small
\caption{Case Existence scoring.}
\label{tab:app-case-e}
\begin{tabular}{lc}
\toprule
Resolution status & Score \\
\midrule
matched & 1.0 \\
excluded & excluded from Trust \\
unresolved, unmatched, empty, or other & 0.0 \\
\bottomrule
\end{tabular}
\end{table}

\begin{table}[t]
\centering
\small
\caption{Case Fidelity and Applicability label-to-score mapping.}
\label{tab:app-case-fa}
\begin{tabular}{p{0.54\columnwidth}c}
\toprule
Label & Score \\
\midrule
\multicolumn{2}{l}{\textit{Case Fidelity}} \\
Consistent & 1.0 \\
Minor omission & 1.0 \\
Key missing content & 0.5 \\
Contradicts & 0.0 \\
Irrelevant & 0.0 \\
Skipped & NA, as 1.0 \\
Other unknown label & 0.0 \\
\midrule
\multicolumn{2}{l}{\textit{Case Applicability}} \\
Supports & 1.0 \\
Partially supports & 0.5 \\
Overextended inference & 0.5 \\
Contradicts & 0.0 \\
Irrelevant & 0.0 \\
Skipped & NA, as 1.0 \\
Other unknown label & 0.0 \\
\bottomrule
\end{tabular}
\end{table}

Case F evaluates whether the report faithfully describes case facts, judicial reasoning, or relevant context. Case A evaluates whether the case applies to and supports the report's local claim.

\subsubsection{NA, Skip, Exclusion, and Unknown Labels}
NA or skipped F/A labels indicate that a Fidelity or Applicability layer has no valid judgment, for example due to insufficient input or an explicit skip. That layer is treated as 1.0 in the multiplicative item score. Excluded items are outside the Trust denominator and do not enter E/F/A averages. Unknown labels are labels not covered by the score mapping; to avoid inflating scores, they are treated as 0.0.

\subsubsection{Report-Level Trust Aggregation}
Report-level Trust pools all included law and case citation items in the report:
\[
\mathrm{Trust}_{\mathrm{report}}=\mathrm{mean}(\mathrm{included\ item\_trust}).
\]
This pooled item average naturally weights law and case citation items by their counts in the report. To avoid hiding type-specific error patterns, the result tables also report law and case E/F/A average scores separately.

If a report has no included law/case citation item, we set $\mathrm{Trust}_{\mathrm{report}}=1.0$. This design treats Trust as conditional citation reliability: it measures reliability given the citations a report includes. Sparse or missing citation behavior is instead reflected by Coverage, Support, Subtopic Evidence Support, citation-count diagnostics, and the Final score.

\subsection{Final Score and System-Level Aggregation}
For each report, the Final score is
\[
\mathrm{Final}=(\mathrm{Coverage}+\mathrm{Support})\times \mathrm{Trust}.
\]
The formula combines report sufficiency with citation trustworthiness. Coverage and Support prevent a system from obtaining a superficially high Trust score by producing very short or citation-sparse reports. Trust penalizes reports whose apparent sufficiency rests on unreliable citations.

System-level leaderboard scores use report-level macro averages. For a system $s$ evaluated on $N_{\mathrm{report}}$ tasks with report-level scores $\mathrm{score}_r$,
\[
\mathrm{SystemScore}_s=\frac{\sum_r \mathrm{score}_r}{N_{\mathrm{report}}}.
\]
This aggregation rule applies to Coverage, Rubric Evidence Support, Subtopic Evidence Support, Support, Trust, and Final. Each report has equal weight in the leaderboard, so a system does not receive extra leaderboard weight merely because it generates more citation items in some tasks.

The dedicated item-level E/F/A diagnostic tables use citation-level micro averages. Columns such as law E, law F, law A, case E, case F, and case A are numeric score averages over citation items of the corresponding type, not binary pass rates. Because E/F/A scoring includes partial-credit labels such as 0.5, these diagnostic columns should be interpreted as mean scores. This item-level view is intended for diagnosing the average reliability of generated law and case references, while the main leaderboard remains report-level.

\section{Robustness of System-Level Trend Analysis}
\label{app:trend-robustness}
The system-level scatter plots in Section~6 combine the main-leaderboard systems with non-duplicate tool-ablation settings and exclude verification-ablation settings. They also exclude Farui. Farui remains in the main leaderboard because it represents a deployed legal product and an end-to-end legal research system. However, in the current tasks, Farui has substantially lower Coverage, Support, and citation count than the other systems. Section~6 focuses on system behavior after systems produce relatively complete long-form reports; excluding Farui from the main scatter plots avoids mixing low-coverage outputs with fuller reports in the same trend analysis.

To check whether this choice changes the main trends, Table~\ref{tab:app-farui-robustness} recomputes Spearman's $\rho$ after including Farui and compares the results with the main plotting setting. As in Section~6, these correlations are descriptive summaries rather than formal significance tests.

\begin{table*}[t]
\centering
\small
\caption{Farui-included robustness check for system-level Spearman correlations.}
\label{tab:app-farui-robustness}
\begin{tabular}{lcccrrrr}
\toprule
Relationship & $x$ & $y$ & Excluded & Main $n$ & Main $\rho$ & With Farui $n$ & With Farui $\rho$ \\
\midrule
Support--Trust & Support & Trust & -- & 9 & -0.9167 & 10 & -0.9394 \\
Coverage--Trust & Coverage & Trust & -- & 9 & 0.6833 & 10 & 0.2242 \\
Coverage--Trust & Coverage & Trust & GPT5 & 8 & 0.5476 & 9 & 0.0833 \\
$n_{\mathrm{ref}}$--Trust & Mean refs. & Trust & -- & 9 & -0.9167 & 10 & -0.9273 \\
Rubric Support--Trust & Rubric Sup. & Trust & -- & 9 & -0.7667 & 10 & -0.8303 \\
Subtopic Support--Trust & Subtopic Sup. & Trust & -- & 9 & -0.9167 & 10 & -0.9394 \\
\bottomrule
\end{tabular}
\end{table*}

Including Farui substantially weakens the Coverage--Trust association. This is expected: Farui is a boundary case with low Coverage, low Support, low citation density, and high Trust in the current tasks. Placing it in the same Coverage--Trust scatter plot as fuller report-producing systems changes the system-level shape of that relationship.

By contrast, Support--Trust, reference count--Trust, Rubric Support--Trust, and Subtopic Support--Trust remain strongly negative after including Farui. Thus, the main findings about the tension among evidence richness, citation density, and Trust do not depend on excluding Farui. The exclusion is intended to keep the behavioral comparison coherent, not to manufacture the Support--Trust tension.

\section{Evaluation Reliability}
\label{app:evaluation-reliability}
The Trust metric relies on automatic citation extraction, source matching, and LLM-based E/F/A judging. To check that the main findings are not artifacts of an unstable evaluation process, we conduct three reliability checks: human audit of sampled E/F/A judgments, repeated-judge stability checks, and report-level bootstrap resampling. These analyses are used to assess reliability and robustness, not as formal significance tests.

Before the final evaluation, we developed the E/F/A protocol through expert-guided pilot auditing. Legal research professionals from a collaborating organization's legal research unit contributed to protocol consultation and audit feedback to varying degrees. The participating experts had graduate-level legal training or equivalent professional background and more than three years of legal-industry experience. One primary expert led most of the protocol refinement and produced the final human-audit labels used in C.1; other members contributed mainly through discussion and consultation. A second expert, with less project-specific involvement than the primary expert, independently audited a subset for the stability check in C.2.2. The experts were not publicly recruited and did not receive per-item or crowdwork-style compensation; participation occurred under an internal collaboration arrangement. They agreed that audit feedback and labels could be used for research evaluation and reported in anonymized, aggregate form. No formal ethics review board approval or exemption determination was obtained; we treated the expert audit as an internal expert-consultation and validation activity rather than a publicly recruited human-subjects study. After an initial demo stage, the research team provided draft annotation instructions to the experts; the experts revised the instructions and clarified legal boundary cases. We then used a small pilot set to iterate on the audit workflow, prompts, model settings, and decision rules. This pilot set was separate from the later human-audit validation sample reported below. The final reported evaluation was run only after the protocol was frozen. The human-audit results should therefore be interpreted as agreement with the expert-guided annotation protocol, not as proof of objective legal truth or as manual correction of final system scores.

\subsection{Human Audit of E/F/A Judgments}
We manually audit sampled E/F/A judgments. During audit, the primary expert auditor reviews the citation context, local claim, and source content, and assigns labels according to the Existence, Fidelity, and Applicability definitions used in this paper. We then compare the LLM judge labels with the primary expert labels at both label level and score level.

Label agreement measures exact agreement on the discrete labels. Score agreement first maps labels to numeric scores and then compares the resulting values. Because some labels share the same score, such as ``minor omission'' and ``consistent'' in Fidelity, score-level agreement is often higher than label-level agreement. This means that some boundary disagreements have limited effect on the final numeric score.

\begin{table*}[t]
\centering
\small
\caption{Human audit of E/F/A judgments.}
\label{tab:app-human-audit}
\begin{tabular}{lrrrrr}
\toprule
Layer & $n$ & Label $\kappa$ & Score $\kappa$ & Label match & Score match \\
\midrule
law E & 257 & 0.8384 & 0.7306 & 0.9883 & 0.9883 \\
law F & 137 & 0.8352 & 0.9182 & 0.8759 & 0.9124 \\
law A & 204 & 0.7970 & 0.9510 & 0.8627 & 0.9455 \\
case E & 107 & 0.7296 & 0.6993 & 0.8692 & 0.8692 \\
case F & 65 & 0.7988 & 0.9555 & 0.8769 & 0.9692 \\
case A & 67 & 0.9206 & 0.9728 & 0.9552 & 0.9552 \\
law macro & 598 & 0.8235 & 0.8666 & 0.9090 & 0.9487 \\
case macro & 239 & 0.8163 & 0.8759 & 0.9004 & 0.9312 \\
overall macro & 837 & 0.8199 & 0.8712 & 0.9047 & 0.9400 \\
\bottomrule
\end{tabular}
\end{table*}

Overall, human-vs-judge agreement is high: overall macro label $\kappa=0.8199$ and score $\kappa=0.8712$. The law and case macro results are also close, suggesting that the E/F/A judge is not stable only for one citation type.

\subsection{Judge and Label Stability}
\subsubsection{LLM Judge Round 1 vs. Round 2}
We further check consistency between two LLM-judge runs on the same validation items. The reproducible validation subset covers law F and law A. Law E is not included in this check because law Existence is produced by deterministic parsing and database verification rather than by the LLM judge; with the same input and database state, this layer is deterministic. This subsection therefore evaluates the stability of semantic law F/A judging rather than human-vs-judge agreement.

\begin{table*}[t]
\centering
\small
\caption{Stability checks for E/F/A judging and expert labels.}
\label{tab:app-stability}
\begin{tabular}{llrrrrr}
\toprule
Check & Layer & $n$ & Label $\kappa$ & Score $\kappa$ & Label match & Score match \\
\midrule
Judge R1 vs. R2 & law F & 137 & 0.8648 & 0.9581 & 0.8978 & 0.9562 \\
Judge R1 vs. R2 & law A & 204 & 0.8460 & 0.9466 & 0.8971 & 0.9412 \\
Judge R1 vs. R2 & law macro & 341 & 0.8554 & 0.9524 & 0.8975 & 0.9487 \\
\midrule
Primary vs. Secondary Expert & law E & 257 & 1.0000 & 1.0000 & 1.0000 & 1.0000 \\
Primary vs. Secondary Expert & law F & 137 & 0.7315 & 0.8391 & 0.7956 & 0.8248 \\
Primary vs. Secondary Expert & law A & 204 & 0.8292 & 0.9578 & 0.8824 & 0.9554 \\
Primary vs. Secondary Expert & law macro & 598 & 0.8536 & 0.9323 & 0.8927 & 0.9267 \\
\bottomrule
\end{tabular}
\end{table*}

The two LLM-judge rounds show high score-level agreement, with law macro score $\kappa=0.9524$. This indicates that although F/A judgments contain semantic boundary cases, the numeric scores used by the benchmark are stable across repeated judging.

\subsubsection{Primary vs. Secondary Expert Audit}
To understand the semantic boundary of the E/F/A annotation task itself, we also compare the primary expert's law E/F/A labels with an independent secondary expert audit on the same subset. The primary expert's labels are treated as the human-audit labels used in C.1; the secondary audit is used only as a stability check, not as a merged label source or manual correction layer. Law E is retained in this check because it is a human database-based existence verification layer rather than an LLM semantic judgment. Its expert-to-expert agreement is therefore 1.0. Law F and law A depend more on contextual interpretation and show a small number of boundary disagreements.

Overall, the expert-label stability results show that Existence is the most deterministic layer, while Fidelity and Applicability involve more interpretation. However, law macro score $\kappa=0.9323$, indicating that even when fine-grained labels differ, score-level judgments remain stable. This supports treating E/F/A as an operational evaluation protocol rather than as a claim that all legal interpretation boundaries can be eliminated.

The difference between C.2.1 and C.2.2 is therefore intentional: C.2.1 evaluates repeated LLM judging and excludes law E because law E does not pass through the LLM judge; C.2.2 evaluates expert-to-expert label stability and retains law E as a deterministic reference layer.

\subsection{Report-Level Bootstrap Robustness}
We conduct bootstrap resampling over report-level scores to check whether the main leaderboard, tool ablation, and verification ablation means are highly sensitive to task sampling. Bootstrap samples reports, not citation items, so that the resampling unit matches the report-level macro averages used in the main leaderboard.

For each system, we resample the 72 reports with replacement 10,000 times and compute the mean score for each sample. Tables~\ref{tab:app-bootstrap-main}--\ref{tab:app-bootstrap-verification} report the observed Trust and Final means, 95\% bootstrap confidence intervals, and bootstrap standard deviations. These results describe stability of mean estimates and are not used as formal significance tests.

\begin{table*}[t]
\centering
\small
\caption{Report-level bootstrap robustness for the main leaderboard.}
\label{tab:app-bootstrap-main}
\begin{tabular}{lrrrrrrr}
\toprule
Model & $n$ & Trust & Trust CI & Trust SD & Final & Final CI & Final SD \\
\midrule
Deli\_DR & 72 & 0.8064 & [0.7644, 0.8469] & 0.0209 & 0.9818 & [0.9237, 1.0398] & 0.0296 \\
GPT5 & 72 & 0.8575 & [0.7883, 0.9182] & 0.0332 & 0.9776 & [0.8975, 1.0512] & 0.0388 \\
Qwen36\_plus & 72 & 0.7744 & [0.7376, 0.8091] & 0.0182 & 0.9612 & [0.9031, 1.0195] & 0.0298 \\
DeepSeek & 72 & 0.8077 & [0.7594, 0.8537] & 0.0240 & 0.9591 & [0.8950, 1.0222] & 0.0325 \\
Deli\_LegalAI & 72 & 0.8041 & [0.7582, 0.8448] & 0.0221 & 0.9221 & [0.8609, 0.9807] & 0.0302 \\
Qwen\_DR & 72 & 0.7164 & [0.6661, 0.7659] & 0.0258 & 0.8856 & [0.8102, 0.9599] & 0.0381 \\
Farui & 72 & 0.9133 & [0.8651, 0.9530] & 0.0223 & 0.3835 & [0.3547, 0.4127] & 0.0148 \\
\bottomrule
\end{tabular}
\end{table*}

The main-leaderboard bootstrap results do not show extreme instability in Trust or Final means. Farui again has high Trust but much lower Final, consistent with the main-paper argument that Trust alone is insufficient for ranking long-form legal research reports.

\begin{table*}[t]
\centering
\small
\caption{Report-level bootstrap robustness for the tool ablation.}
\label{tab:app-bootstrap-tool}
\begin{tabular}{lrrrrrrr}
\toprule
Model & $n$ & Trust & Trust CI & Trust SD & Final & Final CI & Final SD \\
\midrule
Qwen36\_plus & 72 & 0.7744 & [0.7379, 0.8099] & 0.0185 & 0.9612 & [0.9028, 1.0188] & 0.0298 \\
Qwen36\_tool\_law & 72 & 0.7951 & [0.7556, 0.8317] & 0.0193 & 0.9718 & [0.9140, 1.0305] & 0.0298 \\
Qwen36\_tool\_law\_case & 72 & 0.7433 & [0.7128, 0.7727] & 0.0153 & 1.0329 & [0.9792, 1.0882] & 0.0280 \\
Qwen36\_tool\_law\_case\_web & 72 & 0.7441 & [0.7004, 0.7854] & 0.0217 & 0.9903 & [0.9242, 1.0526] & 0.0329 \\
\bottomrule
\end{tabular}
\end{table*}

The tool-ablation bootstrap results match the main-paper trend: adding law and case search gives the highest Final mean, while its Trust mean is lower than the base and law-only settings. This supports the interpretation that tools reshape the Support--Trust trade-off rather than monotonically improving all dimensions.

\begin{table*}[t]
\centering
\small
\caption{Report-level bootstrap robustness for the verification ablation.}
\label{tab:app-bootstrap-verification}
\begin{tabular}{lrrrrrrr}
\toprule
Model & $n$ & Trust & Trust CI & Trust SD & Final & Final CI & Final SD \\
\midrule
Base & 72 & 0.7433 & [0.7127, 0.7727] & 0.0153 & 1.0329 & [0.9786, 1.0864] & 0.0277 \\
E-Filter & 72 & 0.8134 & [0.7839, 0.8416] & 0.0148 & 1.0970 & [1.0501, 1.1435] & 0.0241 \\
EFA-Revise & 72 & 0.8883 & [0.8631, 0.9120] & 0.0124 & 1.1708 & [1.1252, 1.2190] & 0.0240 \\
\bottomrule
\end{tabular}
\end{table*}

The verification-ablation bootstrap results most directly support the conclusion about E/F/A feedback. From Base to E-Filter and then to EFA-Revise, both Trust and Final means increase steadily. This suggests that E/F/A verification feedback is useful not only as a post-hoc diagnostic signal, but also as feedback for a revision stage.

\section{Qualitative Error Analysis}
\label{sec:qualitative-error-analysis}
To complement the quantitative results, we examine three representative citation failures from model-generated reports. These examples are not intended to exhaust all error modes or provide full provenance studies. Instead, they illustrate why legal citation trustworthiness cannot be reduced to whether a report contains citations, or even whether a cited source exists. The three examples correspond to Existence, Fidelity, and Applicability failures.

\subsection{Existence Error}
The first example comes from a Deli\_DR report. The report cites the ``Public Data Management Regulation'' as legal authority. The title has the surface form of a formal regulation, but the verification pipeline cannot resolve it to a valid statute, judicial interpretation, regulation, or normative legal document. The failure therefore occurs before any question of interpretation or legal application: the cited authority itself is not verifiable.

\begin{table}[h]
\centering
\small
\begin{tabular}{p{0.28\columnwidth}p{0.62\columnwidth}}
\toprule
Field & Content \\
\midrule
Model citation & Public Data Management Regulation \\
Verification finding & No valid legal or normative source can be resolved from this title. \\
E/F/A outcome & Existence failure: the cited authority is not verifiable. \\
\bottomrule
\end{tabular}
\caption{Representative Existence failure.}
\label{tab:existence-error}
\end{table}

This is the most basic form of citation hallucination. The system does not merely misread a real source; it introduces a source that appears official but cannot be verified. In a legal research report, such a citation can create a false sense of normative support, because subsequent analysis may appear grounded while relying on an authority that does not exist in the verification sources.

\subsection{Fidelity Error}
The second example comes from a Qwen\_DR report. The report cites a real appellate judgment, case No. (2021) Yue 06 Min Zhong 16937, involving a family property arrangement and a dispute over whether a land-use-right transfer clause should be treated as a gift contract. The report states that the court regarded the arrangement as a gift contract rather than an internal marital property agreement. Because the case itself exists and the underlying facts involve family property arrangements, this is not an Existence failure.

\begin{table}[h]
\centering
\small
\begin{tabular}{p{0.28\columnwidth}p{0.62\columnwidth}}
\toprule
Field & Content \\
\midrule
Model statement & The court treated the land-use-right arrangement as a gift contract rather than an internal marital property agreement. \\
Source finding & ``Zhu Jinhui claims that Clause 3 of the agreement is a gift contract \ldots{} because Clause 3 is one term of the agreement, it \textbf{cannot be separately understood as a gift contract}; therefore, the claim to revoke the gift lacks foundation and is \textbf{rejected}.'' \\
E/F/A outcome & Fidelity failure: the report turns a rejected litigant argument into the court's conclusion. \\
\bottomrule
\end{tabular}
\caption{Representative Fidelity failure.}
\label{tab:fidelity-error}
\end{table}

The problem is that the report reverses the judgment's reasoning. The phrase indicating a ``gift contract dispute'' describes the dispute type, not the court's holding that the specific clause was itself a gift contract. The gift-contract characterization did appear in the case, but only as a party's litigation position. The court's reasoning went in the opposite direction: the agreement contained interrelated family property arrangements, the land-use-right clause could not be isolated from the overall agreement, and the disposition lacked the gratuitous character required for a gift contract. The court therefore rejected the claim to revoke the gift.

This example shows why Fidelity matters even when a citation is real. A reader who checks only whether the case exists may conclude that the report is grounded in a valid authority. Yet the report has extracted a position rejected by the court and rewritten it as the court's rule. A real case has thereby been converted into authority for the opposite legal proposition.

\subsection{Applicability Error}
The third example also comes from a Qwen\_DR report. The report claims that a real-estate gift clause in a divorce agreement has a status-relationship character and is therefore not governed by the general right to revoke a gift. To support this claim, the report cites Article 186 of the Contract Law of the People's Republic of China. The cited article provides that a donor may revoke a gift before transfer of the gifted property right, subject to exceptions for public-interest, moral-obligation, or notarized gifts.

\begin{table}[h]
\centering
\small
\begin{tabular}{p{0.28\columnwidth}p{0.62\columnwidth}}
\toprule
Field & Content \\
\midrule
Local claim & A real-estate gift clause in a divorce agreement is status-related and is not subject to the ordinary revocation rule for gifts. \\
Cited rule & Article 186 states the general pre-transfer revocation rule for gifts and lists limited exceptions. \\
E/F/A outcome & Applicability failure: the rule is real, but it does not establish the status-relationship exception asserted by the report. \\
\bottomrule
\end{tabular}
\caption{Representative Applicability failure.}
\label{tab:applicability-error}
\end{table}

The error is not that the statute is nonexistent, nor primarily that the report misquotes the article. The error is that the article does not support the local claim. Article 186 sets out a general rule permitting revocation before transfer, together with specific exceptions. It does not explain how gifts embedded in divorce agreements should be characterized, nor does it state that a status-relationship agreement categorically excludes the donor's revocation right. If read in isolation, the cited rule more directly supports the opposite proposition: gifts are generally revocable before transfer unless an exception applies.

This example captures an Applicability failure. The citation is real and its text can be identified, but it does not perform the argumentative function assigned to it in the report. Such failures are difficult for citation-existence benchmarks to capture: the source exists, but the legal claim remains unsupported.

\section{Item-Level E/F/A Diagnostics}
\label{app:item-diagnostics}

\begin{table*}[h]
\centering
\scriptsize
\setlength{\tabcolsep}{4pt}
\caption{Main leaderboard item-level E/F/A diagnostics. Scores are per-item averages over generated law and case references.}
\label{tab:main-item-diagnostics}
\begin{tabular}{l r r r r r r r r}
\toprule
Model & $n_{\mathrm{law}}$ & $n_{\mathrm{case}}$ & law E & law F & law A & case E & case F & case A \\
\midrule
Qwen\_DR & 1223 & 158 & 0.9191 & 0.9558 & 0.8814 & 0.7658 & 0.8006 & 0.8291 \\
Qwen36\_plus & 837 & 4 & 0.9104 & 0.9600 & 0.8967 & 0.7500 & 0.6250 & 1.0000 \\
Deli\_DR & 797 & 22 & 0.9341 & 0.9573 & 0.9216 & 0.9091 & 0.7500 & 0.7727 \\
DeepSeek & 603 & 6 & 0.9461 & 0.9552 & 0.8889 & 0.5000 & 0.7500 & 0.5000 \\
Deli\_LegalAI & 564 & 8 & 0.9521 & 0.9495 & 0.9043 & 0.6250 & 0.9375 & 0.8125 \\
Farui & 360 & 0 & 0.6569 & 0.9750 & 0.9597 & -- & -- & -- \\
GPT5 & 175 & 1 & 0.9400 & 0.9543 & 0.8971 & 0.0000 & 1.0000 & 1.0000 \\
\bottomrule
\end{tabular}
\end{table*}

\begin{table*}[h]
\centering
\scriptsize
\setlength{\tabcolsep}{4pt}
\caption{Tool ablation item-level E/F/A diagnostics. Scores are per-item averages over generated law and case references.}
\label{tab:tool-item-diagnostics}
\begin{tabular}{l r r r r r r r r}
\toprule
Model & $n_{\mathrm{law}}$ & $n_{\mathrm{case}}$ & law E & law F & law A & case E & case F & case A \\
\midrule
Qwen36\_plus & 837 & 4 & 0.9104 & 0.9600 & 0.8967 & 0.7500 & 0.6250 & 1.0000 \\
Qwen36\_tool\_law & 805 & 2 & 0.9453 & 0.9503 & 0.8826 & 0.0000 & 1.0000 & 1.0000 \\
Qwen36\_tool\_law\_case & 849 & 146 & 0.9446 & 0.9441 & 0.8810 & 0.8014 & 0.8116 & 0.8288 \\
Qwen36\_tool\_law\_case\_web & 814 & 73 & 0.9244 & 0.9595 & 0.8839 & 0.6575 & 0.8562 & 0.8767 \\
\bottomrule
\end{tabular}
\end{table*}

\begin{table*}[h]
\centering
\scriptsize
\setlength{\tabcolsep}{4pt}
\caption{Verification ablation item-level E/F/A diagnostics. Scores are per-item averages over generated law and case references.}
\label{tab:verification-item-diagnostics}
\begin{tabular}{l r r r r r r r r}
\toprule
Model & $n_{\mathrm{law}}$ & $n_{\mathrm{case}}$ & law E & law F & law A & case E & case F & case A \\
\midrule
Base & 849 & 146 & 0.9446 & 0.9441 & 0.8810 & 0.8014 & 0.8116 & 0.8288 \\
E-Filter & 751 & 104 & 0.9987 & 0.9434 & 0.8775 & 0.9808 & 0.7837 & 0.7644 \\
EFA-Revise & 748 & 74 & 0.9866 & 0.9739 & 0.9265 & 0.9595 & 0.9257 & 0.9527 \\
\bottomrule
\end{tabular}
\end{table*}

This appendix reports the citation-level counterparts of the report-level summaries in Section~5. The main tables compare systems at report level: each task contributes one report-level score, and E/F/A columns are first aggregated within reports and then macro-averaged. The tables below instead use citation-level micro averages over generated law and case references. They are not leaderboard scores. Their purpose is diagnostic: they expose how many legal references each system produced and how reliable those concrete references were on average.

This distinction matters because citation trustworthiness has two faces. A report-level Trust score asks how reliable a system is across tasks. An item-level diagnostic table asks what happens on the surface where legal authority is actually used: individual statutes, regulations, judicial interpretations, and cases. A system can improve Support by citing more authorities while also expanding the number of opportunities for Existence, Fidelity, or Applicability failures. Conversely, a system can look highly trustworthy if it cites little, which is why citation counts and Support remain necessary companions to Trust.

Table~\ref{tab:main-item-diagnostics} shows three patterns behind the main leaderboard. First, most systems generate many more law citations than case citations, so case-level E/F/A scores should be read together with $n_{\mathrm{case}}$; a perfect case score over one or two items is not comparable to a lower score over more than one hundred cases. Second, law Fidelity is generally high once a source is resolved, suggesting that many systems can paraphrase or describe statutory material reasonably well after locating it. Larger differences appear in law Existence and Applicability, where retrieval coverage, source selection, and claim-source matching matter more. Third, Qwen\_DR generates the largest number of both law and case references. This helps explain its strong Support, but it also increases exposure to citation errors, especially on case citations where source structure and factual fit are harder to verify than statute names and articles.

Table~\ref{tab:tool-item-diagnostics} clarifies why retrieval tools improve evidence support without monotonically improving Trust. Adding law search does not increase citation volume much, and its law Existence score improves relative to the base setting. Adding case search changes the system more sharply: case citations increase from 4 to 146, which explains the stronger Subtopic Evidence Support in the report-level table but also creates a larger case-level verification burden. Adding web search reduces case citations to 73 and does not improve Trust, suggesting that broader web access changes evidence selection rather than reliably making legal citations more trustworthy. The lesson is not that case search is harmful; it is that legal retrieval creates candidate authority, while citation trustworthiness still depends on selecting the right authority and using it for the right local claim.

Table~\ref{tab:verification-item-diagnostics} shows where the verification-ablation gains come from. E-Filter behaves like an existence-level cleanup stage: law E rises from 0.9446 to 0.9987, and case E rises from 0.8014 to 0.9808. However, case Fidelity and Applicability decrease after filtering, meaning that removing unresolved citations does not make the remaining case uses faithful or locally supportive. EFA-Revise has a different profile. It keeps Existence high while improving the semantic dimensions more clearly: law F, law A, case F, and case A all improve relative to Base. This is the key diagnostic reason that full E/F/A feedback improves Final more clearly than existence-only filtering. The useful signal is not only ``delete citations that do not exist,'' but also ``revise or remove citations whose described content or argumentative role does not match the source.''

Taken together, the item-level diagnostics explain the central tension in LegalCiteTrust. Support rewards legal evidence coverage, but more evidence increases the number of citation-level commitments a report makes. Trust penalizes failures in those commitments, but Trust alone cannot distinguish a careful complete report from a sparse report that avoids difficult authority. The paired report-level and item-level views therefore work together: report-level scores compare task performance, while item-level diagnostics reveal whether a system's evidence behavior is built on reliable legal references or on a larger surface of fragile citation use.

\section{Additional Reproducibility and Responsible-Use Details}
\label{app:additional-reproducibility}

\subsection{Computational and Implementation Details}
\label{app:computational-implementation-details}

Most evaluated systems are closed or product-level services, so parameter counts and provider-side compute are not publicly disclosed. We therefore report system identifiers, access conditions, and default generation settings rather than parameter sizes.

The computational budget consists of report generation through provider APIs or product interfaces and citation-verification evaluation. Provider-side generation compute and product-internal retrieval costs are not observable for closed systems. For the evaluation pipeline, a single 72-report benchmark run costs roughly 10 RMB and takes about 2 hours in our setup, excluding earlier development, prompt tuning, debugging, and validation runs. This estimate is approximate because runtime and cost vary with the number and type of citations in generated reports.

The released code records the task prompts, citation extraction, source resolution, E/F/A judging, scoring, and aggregation pipeline. Local computation is used for orchestration, parsing, scoring, aggregation, and bootstrap analysis, while model inference and product-level retrieval are performed through the corresponding provider services.

\subsection{Societal Impact and Risks}
\label{app:societal-risks}

LegalCiteTrust is intended as a research benchmark for evaluating citation trustworthiness in Chinese long-form legal research reports, not as legal advice, a certification of legal correctness, or a final adjudication of legal documents. Although the Existence, Fidelity, and Applicability framework is designed to be general, its category boundaries, operational definitions, and judge prompts are necessarily shaped by the researchers' and legal consultants' expertise, assumptions, and available effort. The human-audit results should therefore be interpreted as agreement with the study's annotation protocol rather than as proof of agreement with objective legal truth. A potential misuse risk is that automated E/F/A scores could be overinterpreted in serious legal settings as replacing professional legal review. LegalCiteTrust should instead be used for preliminary screening, diagnosis, and research comparison, with further expert validation required before any high-stakes deployment.

\subsection{AI Assistant Use}
\label{app:ai-assistant-use}

The authors used Codex as an AI assistant during manuscript preparation and release preparation. Its use included language editing, LaTeX editing, checklist drafting, release-documentation support, and code/documentation inspection. Codex was also used to help identify inconsistencies across manuscript text, checklist responses, and repository documentation. It was not used as an autonomous author, a source of legal authority, or a source of unverified scientific claims. All scientific claims, experimental results, legal interpretations, code changes, and release decisions were reviewed and approved by the authors, who take responsibility for the final manuscript and artifacts.


\end{document}